\documentclass[conference]{IEEEtran}
\usepackage{times}

\usepackage[numbers]{natbib}
\usepackage{multicol}
\usepackage[bookmarks=true]{hyperref}
\usepackage{amssymb}
\usepackage{graphicx}
\usepackage{xcolor}
\usepackage{pgfplots}

\begin{document}

\title{Language-Conditioned Change-point Detection to Identify Sub-Tasks in Robotics Domains}

\author{Author Names Omitted for Anonymous Review. Paper-ID [add your ID here]}
\author{Divyanshu Raj, Chitta Baral, Nakul Gopalan\\
School of Computing and Augmented Intelligence, Arizona State University\\
\{draj5, cbaral, ng\}@asu.edu}



%

\newcommand{\mycomment}[1]{}

\maketitle

\begin{abstract}

In this work, we present an approach to identify sub-tasks within a demonstrated robot trajectory using language instructions. 
We identify these sub-tasks using language provided during demonstrations as a guidance to identify sub-segments of a longer robot trajectory.
Given a sequence of natural language instructions and a long trajectory consisting of image frames and discrete actions, we want to map an instruction to a smaller fragment of the trajectory.
Unlike previous instruction following works which directly learn the mapping from language to a policy, we propose a language-conditioned change-point detection method to identify sub-tasks in a problem. Our approach learns the relationship between constituent segments of a long language command and corresponding constituent segments of a trajectory. 
These constituent trajectory segments can be used to learn sub-tasks or sub-goals for planning or options as demonstrated by previous related work~\cite{ngpln}. 
Our insight in this work is that the language-conditioned robot change-point detection problem is similar to the existing video moment retrieval works~\cite{lei2020tvr,lei2021qvhighlights} used to identify sub-segments within online videos.
Through extensive experimentation, we demonstrate a $1.78_{\pm 0.82}\%$
 improvement over a  baseline approach in accurately identifying sub-tasks within a trajectory using our proposed method. 
 Moreover, we present a comprehensive study aimed at investigating sample complexity requirements on learning this mapping, between language and trajectory sub-segments, to understand if the video retrieval based methods are realistic in real robot scenarios.    
 
\mycomment{
\begin{itemize}
    \item We are trying to identify sub-tasks within a demonstrated trajectory given language instructions.
    \item Previous works learn mapping from language to a policy directly
    \item Our Change point detection method  learns a relationship between parts of a longer command and parts of a trajectory that satisfies a that segment of a command
    \item we demonstrate x\% improvement over a baseline approach to do this task.
\end{itemize}
}
\end{abstract}

\IEEEpeerreviewmaketitle

\section{Introduction}

Language-to-policy grounding approaches generally map language directly to policies without explicitly learning which parts of the language commands map to which constituent sub-tasks of a demonstrated trajectory. 
Such an approach can cause difficulty in reusing sub-tasks to solve novel tasks even if these sub-tasks exist within previously demonstrated tasks. 
We present our approach for identifying sub-tasks within a demonstrated trajectory using paired language instructions.
Our approach breaks down instructions into sub-segments and trajectories into corresponding sub-tasks. Natural language instructions such as “Turn around and go to the desk” are segments of a longer instructions. Their corresponding trajectory consisting of image frames and actions can be considered as demonstrations for their sub-tasks as shown in Figure~\ref{fig:init1o}. Such an approach can allow the robot to  comprehend and identify sub-tasks, improving their capacity to generalize and solve novel longer horizon tasks. 

Instruction-following problems solved using techniques such as Semantic parsing  required pre-specifying goal and subgoal conditions~\cite{32,33}. However, recent approaches such as learning mappings from language to reward functions through inverse reinforcement learning and leveraging large-scale language model pre-training have enabled robots to learn tasks and follow instructions more effectively. These approaches addressed the limitations of traditional methods and improved exploration in reinforcement learning. Another line of work related to our approach is moment retrieval, where a moment of a video is retrieved given a natural language command~\cite{li2020hero}. We approach the robot sub-task identification problem as a video moment retrieval problem~\cite{lei2021qvhighlights}, that is, to identify the part of a robot trajectory representing a sub-task that is specified by a segment of the complete language command. 
Moreover, we have transformed the ALFRED~\cite{ALFRED20} dataset to a changepoint detection dataset, where we combine sequential low-level language instructions as the long language instruction and find the constituent trajectory segment for each constituent segment of language instructions.


Our work makes three significant contributions. Firstly, we successfully adapted the moment retrieval technique, commonly used in video analysis with natural language queries, to the robotics domain. This adaptation enabled us to detect changepoints within robot trajectories based on natural language instructions, resulting in improved changepoint detection compared to previous approaches that learned sub-tasks separately. Additionally, we conducted an ablation study that explored different trajectory definitions and assessed the impact of training data size on the models' ability to accurately localize and segment trajectories. The findings from this study provide valuable insights for future research in this field.

\mycomment{
Through our empirical analysis, we reveal intriguing insights into the relationship between training data size and IoU performance. We observe that as the training data decreases, the IoU scores exhibit a clear downward trend. Our results demonstrate the pronounced influence of training data availability on the ability of models to accurately localize and segment in trajectories.
\begin{itemize}
    \item We are trying to identify sub-tasks within a demonstrated trajectory given language instructions. Where is this useful?
    \item Previous work has used changepoint detection on robots for learning options and for language grounding. We want to improve upon the change point detection techniques on a robot. (Related work in a para)
    \item Para to describe approach
    \item Para to list contributions:
    \begin{itemize}
        \item converting alfred dataset to a changepoint detection dataset
        \item Adapt a video momemnt detection model to use robot video and trajectory data for change point detection
        \item sample efficiency results on our model to see how data starved our model can be to detect changepoints in novel problems
    \end{itemize}
\end{itemize}
}

\section{Related Work}
Semantic parsing has been used to solve instruction-following problems~\cite{32,33}. This requires goal conditions and subgoal conditions to be pre-specified. MacGlashan et al. \cite{34} learned mapping from language to a reward function learned via Inverse Reinforcement learning with objects and rooms  pre-specified in their domains, and not learned from scratch. Other end-to-end learning methods require millions of episodes to learn
simple behaviors using reinforcement learning \cite{37}, \cite{38}.
Gopalan et al. \cite{ngpln} address the challenge of enabling robots to learn tasks and follow instructions by learning sub-tasks using change point detection and then mapping language to these sub-tasks demonstrating generalizable planning. In this work we only attempt to solve the language-based changepoint detection problem at a large scale and do not attempt to use these segments for planning. 
The paper \cite{du2023guiding} introduces ELLM (Exploring with LLMs), a method that utilizes large-scale language model pretraining to shape exploration in reinforcement learning. 

Previous work in video moment retrieval has demonstrated the use of language to retrieve a segment of a video that matches the language query.  \cite{gao2017tall, hendricks2017localizing, lei2021qvhighlights, lei2020tvr}. 
More recent approaches use multi-modal language-based end-to-end transformer models to identify spoken sentences~\cite{yuan2018talk} or moments within videos~\cite{cao2022pursuit,zhang2019exploiting, rodriguezopazo2020proposalfree, nam2021zeroshot} specified using a language query.  
The paper by Lei et al. \cite{lei2021qvhighlights} introduces the Query-based Video Highlights (QVHIGHLIGHTS) dataset, consisting of over 10,000 YouTube videos annotated with human-written natural language queries, relevant moments, and saliency scores. 

\section{Dataset Creation from ALFRED}
We have used ALFRED \cite{ALFRED20} (Action Learning From Realistic Environments and Directives), a dataset for learning a mapping from natural language instructions to robotics trajectory for household tasks. ALFRED consists of expert demonstrations for 25k natural language directives. These directives contain high-level goals like “Examine the vase by lamplight.” and low-level language instructions like “Turn around and go to the desk”. We have used low-level language instructions, discrete actions, and image frames to make a new dataset for our experiment. Visual example for the above trajectory demonstration is shown in Figure[\ref{fig:init1o}, \ref{fig:init2o}, \ref{fig:init3o}, \ref{fig:init4o}].

\subsection{Conversion Process}
The ALFRED dataset has $2435$ tasks and each task has three trials. For each trial, there are three sets of high-level and low-level language instructions. First, the image frames are converted to videos with a frame rate of five frames per second. The low-level language instructions in ALFRED that were mapped to image frames are now mapped to video seconds as \emph{relevant\_windows}.
The video is further broken down into two-second clips with 0 index and relevant indexes are stored as \emph{relevant\_clip\_ids}. For example, if the low-level language instruction "Walk to the coffee maker on the right" is mapped to image frame numbers from 10 through 50, the \emph{relevant\_windows} with have the window [2, 10] (in seconds) and the \emph{relevant\_clip\_ids} will have [1, 2, 3, 4]. For each entry in \emph{relevant\_clip\_ids}, we have assigned a highlight score of four which is the highest because all the moments are important in the selected window. 
The trajectory also includes discrete actions. We map the actions to a number of frames, as one action can lead to multiple image frames. A discrete action \emph{MoveForward} can refer to multiple image frames. As moving forward may require the robot to move forward multiple times which will change image frames multiple times whereas another action \emph{PickObject} may only refer to one movement resulting in the change of just one image frame. These discrete actions need to be mapped with the image frames for data creation. Then the actions are mapped to the video seconds in a similar way and then reducing it by five times after mapping. At the end of this pre-processing one or multiple discrete actions have been mapped to every second of the video.

\begin{figure}[htbp]
    \centering
    \includegraphics[scale=0.42]{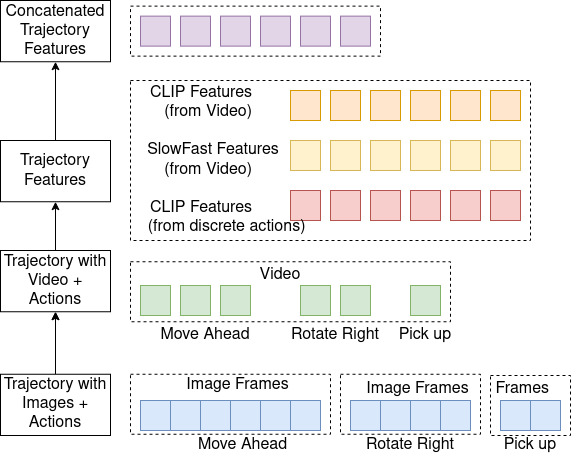}
    \caption{Trajectory Preparation}
    \label{fig:init}
\end{figure}

The final dataset has trajectories with corresponding videos of \emph{n} seconds, and a \emph{n} size discrete action list. For all the low-level language instruction, which is on average six per trajectory, we have a \emph{relevant\_windows}, \emph{relevant\_clip\_ids}, and saliency scores.
The process of preparation of a trajectory is shown in Figure[\ref{fig:init}] -
\begin{table*}[]
\resizebox{\textwidth}{!}{%
\begin{tabular}{ccccccccccccccc}
\multicolumn{3}{c}{Features} &  &  &  &  &  & \multicolumn{5}{c}{Changepoint Detection} & \multicolumn{2}{c}{Highlight Detection} \\ \cline{1-3} \cline{9-10} \cline{11-15}
CLIP & Actions & SlowFast & \multicolumn{5}{c}{Loss} & \multicolumn{2}{c}{R1} & \multicolumn{3}{c}{mAP} & \multicolumn{2}{c}{\textgreater{}= Very Good} \\ \hline
Video & Actions & Video & L1 & gIoU & Saliency & CLS & Contrastive & @0.5 & @0.7 & @0.5 & @0.75 & avg & mAP & HIT@1 \\ \hline
$\checkmark$ &  & $\checkmark$ & $\checkmark$ & $\checkmark$ & $\checkmark$ & $\checkmark$ &  & ${83.6_{\pm{0.47}}}$ & ${64.2_{\pm{0.77}}}$ & ${89.4_{\pm{0.29}}}$ & ${68.8_{\pm{0.72}}}$ & ${65.5_{\pm{0.21}}}$ & ${82.6_{\pm{0.58}}}$ & ${73.7_{\pm{1.00}}}$ \\
$\checkmark$ & $\checkmark$ & $\checkmark$ & $\checkmark$ & $\checkmark$ & $\checkmark$ & $\checkmark$ &  & ${84.8_{\pm{0.39}}}$ & ${65.6_{\pm{0.59}}}$ & ${90.6_{\pm{0.34}}}$ & ${70.3_{\pm{0.46}}}$ & $\mathbf{{66.7_{\pm{0.19}}}}$ & ${85.3_{\pm{0.33}}}$ & ${77.6_{\pm{0.73}}}$ \\
$\checkmark$ & $\checkmark$ & $\checkmark$ & $\checkmark$ & $\checkmark$ & $\checkmark$ & $\checkmark$ & $\checkmark$ & $\mathbf{{85.1_{\pm{0.48}}}}$ & $\mathbf{{65.8_{\pm{0.74}}}}$ & $\mathbf{{90.9_{\pm{0.31}}}}$ & $\mathbf{{70.4_{\pm{0.48}}}}$ & ${66.7_{\pm{0.22}}}$ & $\mathbf{{85.5_{\pm{0.63}}}}$ & $\mathbf{{78.2_{\pm{1.30}}}}$ \\ \hline
\end{tabular}}
\caption{Ablation Study with different definition of trajectories}
\label{tab:my-table-2}
\end{table*}

\subsection{Processed Dataset Format}
The processed dataset contains $6.8$K trajectories in total. There are $146.1$K low-level language instructions across all the trajectories. For training the model, we need features per trajectory and features for every query. This dataset can be further used for various other tasks such as next sub-task prediction, and trajectory retrieval.

We have extracted trajectory features as shown in Figure[\ref{fig:init}]. For each trajectory $t$ we have a video $v$ comprised of a sequence of $L_v$ clips, and discrete action $a$  comprised of a sequence of $L_a$ actions where $L_v = L_a$. The video $v$'s feature denoted by $E_v$ is extracted using HERO~\cite{li2020hero} feature extractor, which extracts CLIP \cite{radford2021learning} features $E_{v(CLIP)} \in \mathcal{R}^{(L_v/2) \times 512}$ and SlowFast \cite{feichtenhofer2019slowfast} features $E_{v(SF)} \in \mathcal{R}^{(L_v/2) \times 2304}$ for every two second clip. The discrete action $a$ whose feature is denoted by $E_a$ is extracted using CLIP features $E_a \in \mathcal{R}^{(L_{a}/2) \times 512}$. Finally, the trajectory $t$ feature denoted by $E_t$ is the concatenation of video features $E_v$ and action features $E_a$, $E_t \in \mathcal{R}^{(L_v/2) \times 3328}$. For the low-level language instruction $q$ of $L_q$ tokens, the feature denoted by $E_q$ is extracted using CLIP features, where $E_q \in \mathcal{R}^{L_q \times 512}$. Table~\ref{tab:my-table} in the Appendix contains the analysis of the new dataset. 
\begin{table*}[]
\resizebox{\textwidth}{!}{%
\begin{tabular}{cccccccccc}
\multicolumn{3}{c}{Dataset} & \multicolumn{5}{c}{Changepoint Detection} & \multicolumn{2}{c}{Highlight Detection} \\ \hline
\multicolumn{2}{c}{Training Set} & Valid Seen & \multicolumn{2}{c}{R1} & \multicolumn{3}{c}{mAP} & \multicolumn{2}{c}{= Very Good} \\ \hline
percentage & total & total & @0.5 & @0.7 & @0.5 & @0.75 & avg & mAP & HIT@1 \\ \hline
100 & 6561 & 250 & $\mathbf{{85.1_{\pm{0.48}}}}$ & $\mathbf{{65.8_{\pm{0.74}}}}$ & $\mathbf{{90.9_{\pm{0.31}}}}$ & $\mathbf{{70.4_{\pm{0.48}}}}$ & $\mathbf{{66.7_{\pm{0.22}}}}$ & $\mathbf{{85.5_{\pm{0.63}}}}$ & $\mathbf{{78.2_{\pm{1.30}}}}$ \\
50 & 3280 & 250 & ${81.8_{\pm{0.92}}}$ & ${62.9_{\pm{0.95}}}$ & ${88.6_{\pm{0.63}}}$ & ${68.3_{\pm{0.77}}}$ & ${64.6_{\pm{0.53}}}$ & ${82.2_{\pm{0.44}}}$ & ${72.3_{\pm{0.97}}}$ \\
25 & 1640 & 250 & ${76.6_{\pm{1.20}}}$ & ${58.5_{\pm{1.32}}}$ & ${84.6_{\pm{0.90}}}$ & ${64.5_{\pm{1.09}}}$ & ${61.2_{\pm{0.79}}}$ & ${77.9_{\pm{0.89}}}$ & ${65.1_{\pm{1.71}}}$ \\
20 & 1312 & 250 & ${73.9_{\pm{1.59}}}$ & ${55.3_{\pm{1.87}}}$ & ${82.4_{\pm{1.26}}}$ & ${61.7_{\pm{1.72}}}$ & ${59.0_{\pm{1.15}}}$ & ${75.6_{\pm{1.01}}}$ & ${62.1_{\pm{1.32}}}$ \\
15 & 984 & 250 & ${71.4_{\pm{2.02}}}$ & ${53.8_{\pm{2.24}}}$ & ${80.6_{\pm{1.52}}}$ & ${60.8_{\pm{1.76}}}$ & ${57.5_{\pm{1.43}}}$ & ${73.0_{\pm{1.61}}}$ & ${58.9_{\pm{1.94}}}$ \\
10 & 656 & 250 & ${66.7_{\pm{1.38}}}$ & ${50.0_{\pm{1.10}}}$ & ${76.6_{\pm{1.33}}}$ & ${57.4_{\pm{1.06}}}$ & ${54.1_{\pm{1.09}}}$ & ${70.5_{\pm{0.70}}}$ & ${55.4_{\pm{1.03}}}$ \\
5 & 328 & 250 & ${57.7_{\pm{1.25}}}$ & ${42.1_{\pm{1.07}}}$ & ${68.6_{\pm{1.43}}}$ & ${49.6_{\pm{1.43}}}$ & ${47.0_{\pm{1.18}}}$ & ${65.5_{\pm{1.10}}}$ & ${50.7_{\pm{1.26}}}$ \\
3 & 197 & 250 & ${47.8_{\pm{1.23}}}$ & ${33.5_{\pm{1.24}}}$ & ${59.1_{\pm{1.25}}}$ & ${40.1_{\pm{1.70}}}$ & ${38.2_{\pm{1.28}}}$ & ${61.4_{\pm{1.05}}}$ & ${45.9_{\pm{1.49}}}$ \\
2 & 131 & 250 & ${40.7_{\pm{2.50}}}$ & ${26.1_{\pm{2.35}}}$ & ${51.8_{\pm{2.62}}}$ & ${30.7_{\pm{2.91}}}$ & ${30.6_{\pm{2.30}}}$ & ${58.5_{\pm{1.10}}}$ & ${42.8_{\pm{1.59}}}$ \\ \hline
\end{tabular}}
\caption{Ablation Study with varying training data size}
\label{tab:my-table-3}
\end{table*}
\section{Methods}
We want to detect the sub-tasks in the trajectories from the given low-level natural language instructions. 
Given a low-level natural language instruction $q$ of $L_q$ tokens, and a trajectory $t$ of a sequence of $L_t$ seconds comprising of video $v$ of $L_v$ clips and discreet actions $a$ of $L_a$ length, where $L_t = L_v = L_a$, we want to find a changepoint in a trajectory which will be signified by consecutive frames from the trajectory, that are a series of images denoted as ${f_i}$. We have taken the transformers-based video localization model used in the QvHighlights \cite{lei2021qvhighlights} for our dataset (converted from ALFRED\cite{ALFRED20}), and in this experiment, we present a strong baseline and ablation study for our dataset. The Moment-DETR model views localization as a direct set prediction, which is similar to our approach for changepoint detection in a trajectory.

\subsection{Architecture}
The input for the transformer encoder, same as in \cite{40}, \cite{3}, \cite{lei2021qvhighlights} is the features of the entire trajectory denoted by $E_t \in \mathcal{R}^{(L_v/2) \times 3328}$, and the features of low-level language instruction denoted by $E_q \in \mathcal{R}^{L_q \times 512}$ as described in the dataset creation process. After that, we use separate 2-layer perceptrons with layernorm \cite{14} and dropout \cite{14} to project the trajectory and low-level natural language instruction features into a shared embedding space of size $d$. The projected features are concatenated and serve as the input to the transformers encoder model, denoted as $E_{input} \in \mathcal{R}^{L \times d}, L=L_t+L_q$.

The trajectory and low-level language instructions as the input sequence are encoded in a stack of the transformer's encoder layers. Each encoder layer has the same architecture with a multi-head self-attention layer and a Feed Forward network. Positional encoding is also added to the input of each attention layer. The output of the encoder is $E_{encoder} \in \mathcal{R}^{L \times d}$. This is passed through a stack of Transformers decoder layers, where each layer consists of a multi-head attention layer, a cross-attention layer that allows self-attention between encoder outputs and decoder inputs, and finally a Feed Forward Network. The decoder input is a set of $N$ trainable positional embeddings of size $d$. The output of the decoder is $E_{decode} \in \mathcal{R}^{N \times d}$.

The salience scores $S \in \mathcal{R}^{L_v}$ are predicted from encoder output $E_{encoder}$ for the given trajectory. The decoder output $E_{decoder}$ is passed through a 3-layer Feed Forward Network with ReLU \cite{11} to predict the normalized changepoint center coordinate and width for the trajectory. We use a linear layer with softmax to predict class labels. In this dataset, since class labels are not available, we assign it as $foreground$ if it matches ground truth or $background$ otherwise as done in \cite{lei2021qvhighlights}.

We used the same loss function as used by \citeauthor{lei2021qvhighlights} \cite{lei2021qvhighlights}  for moment localization, where predictions are matched with ground truth moments using the Hungarian algorithm. The matching cost, quantifies the dissimilarity between a prediction and a ground truth moment, considering the class label (foreground or background) and the changepoint center coordinates and width. The optimal assignment between predictions and ground truth moments is found by minimizing the sum of matching costs. The changepoint detection loss measures the discrepancy between the predicted and ground truth moments using an $L1$ loss and a generalized $IoU$ loss. The saliency loss is computed as a hinge loss between pairs of positive and negative clips within and outside the ground truth moments. 
The saliency/highlightness is assigned on a two-second clip level on a scale of 0-4, 0 meaning "Very Bad" and 4 meaning "Very Good". For our experiment, all the two-second clips have "Very Good" assignments.
The overall loss is a linear combination of the saliency loss, the classification loss, and the moment localization loss, with hyperparameters to balance their contributions. $\lambda_{L1}$, $\lambda_{iou}$, $\lambda_{cls}$, $\lambda_{s}=4$ are hyperparameters for balancing the loss, $IoU$, fore/background, and saliency loss respectively.
We have also used contrastive loss in addition to the above loss functions for experimentation.

\subsection{Feature Selection}
We have defined the trajectory as the combination of image frames and discrete actions. Further, we have transformed the trajectory to a combination of video (from image frames) and discrete actions. 
We have the trajectory $t$ that comprises only the videos $v$ with the trajectory features being the same as the video features, i.e., $E_t = E_v$. Our method accepts a the trajectory $t$ that comprises videos $v$ and discrete actions $a$. The original dataset is such that the length of video clips is not equal to the number of discrete actions $L_v \neq L_a$. The reason for this difference is because a discrete action such as $MoveForward$ can refer to multiple image frames whereas another action such as $PickObject$ may be completed with a single image frame. We mapped the length of discrete actions $L_a$ to the image frames and then pruned it as shown in Figure[\ref{fig:init}], resulting in a trajectory $t$ that has videos $v$ and discrete actions $a$ such that $L_v = L_a$. Finally, we use these two trajectory definitions for experimentation.

\section{Results}
\subsection{Experiment Setup}
We have used the transformed version of the ALFRED dataset, and we use the training data which is split into $90\%$ train and $10\%$ val and \emph{valid seen} as $test$ set data. To evaluate the changepoint detection, we use mean average precision (mAP) with IoU thresholds of $0.5$ and $0.75$, as well as the average mAP over multiple IoU thresholds $[0.5: 0.05: 0.95]$. We have used the standard metric Recall@1 (R@1) used in single changepoint detection, where the prediction is positive for two sets of IoU, $0.5$ and $0.7$. For highlight detection, we used mAP and HIT@1 to compute the hit ratio for the highest-scored clip. The saliency score is assigned as "very good" for all as all the moments in the sub-task in the trajectory are considered important in this dataset.
\begin{tikzpicture}

\begin{axis}[legend style={at={(0.95,0.3)},anchor=north east},
symbolic x coords={0, 2, 3, 5, 10, 15, 20, 25, 50, 100}, xtick=data,
xmin = 0,
ymin = 0, xlabel = {Percentage(\%) of Training data used}]

\addlegendentry{R1@0.5 ($\mu$)}
\addplot[mark=*,thick,blue] coordinates {
(2,40.7)
(3,47.8)
(5,57.7)
(10,66.7)
(15,71.4)
(20,73.9)
(25,76.6)
(50,81.8)
(100,85.1)
};

\addlegendentry{mAP@0.5 ($\mu$)}
\addplot[mark=diamond*,thick,red] coordinates {
(2,51.8)
(3,59.1)
(5,68.6)
(10,76.6)
(15,80.6)
(20,82.4)
(25,84.6)
(50,88.6)
(100,90.9)
};
\end{axis}

\end{tikzpicture}

\mycomment{

\subsection{Implementation Details}
The model is implemented in PyTorch. We set the hidden size $d = 256$, the number of layers in the encoder and decoder layer, $T=2$, and the number of possible changepoint is $N=10$. We use a dropout of $0.1$ for transformer layers and $0.5$ for input projection layers. We set the loss fusion hyper parameters as $\lambda_{L1}=10$, $\lambda_{iou}=1$, $\lambda_{cls}=4$, $\lambda_{s}=1$, $\Delta=0.2$. The model weights are initialized with Xavier init. We use AdamW with an initial learning rate of $1e-4$, and weight decay of $1e-4$ to optimize model parameters. The model is trained for $200$ epochs with a batch size of $256$. All training is conducted on a single NVIDIA GeForce RTX $3090$ GPU, with a training time of $4$ hours for a single iteration.
}

\subsection{Results and Analysis}
We have done two ablation studies to understand how to best detect the changepoint. The first study is shown in Table \ref{tab:my-table-2} in which we have shown the impact of defining our trajectory in different ways. We defined trajectory in two ways, where first the trajectory $t_1$ is just represented by videos $v$ and the trajectory feature is the same as video feature $E_{t1} = E_v \in \mathcal{R}^{(L_v/2) \times 2816}$, and the second trajectory $t_2$ is defined as the combination of videos $v$ and discrete actions $a$, where $L_v = L_a$, trajectory feature $E_{t2} \in \mathcal{R}^{(l_v/2) \times 3328}$. We have used the first trajectory definition $t1$ as our baseline result here. We observe that there has been $2.1\%$ improvement in the mean value for $R1@0.7$ and $2.1\%$ improvement for $mAP@0.75$ for changepoint detection. We also show significant improvement for highlight detection with $3.2\%$ improvement for $mAP$ and $5.2\%$ improvement for $HIT@1$. This shows the significance of discrete actions discussed  in natural language in detecting sub-segments of a trajectory. 
For the second trajectory $t_2$, we further use contrastive loss and observe a slight improvement for all the metrics.

In the second ablation study shown in Table \ref{tab:my-table-3} we show the effect on metrics by decreasing the number of trajectories in training data. In a real-world robotics scenarios, having a large number of sample trajectories can be impossible, so this study is to evaluate how many trajectories would be required on an actual robot. With $2\%$ of training data, we need $131$ trajectory demonstrations, with $R1@0.5$ for changepoint detection of $40.7_{\pm{2.5}}$ and $mAP@0.75$ of $30.7_{\pm{2.9}}$. With $50\%$ of the data, i.e., $3280$ trajectories we get our performance of over $80\%$ in $R1@0.5$. This is more reasonable in terms of performance but still requires a large number of samples indicating an unsolved research problem.

\section{Conclusion}

This work presents an approach to identify sub-tasks within a demonstrated robot trajectory using language instructions. Unlike previous methods that directly map language to policies, the proposed approach employs a language-conditioned change-point detection method. Through extensive experimentation on the transformed ALFRED dataset, the results demonstrate improved sub-task identification accuracy compared to baseline approaches. The adaptation of moment retrieval techniques from videos to the robotics domain proves effective in localizing and segmenting trajectories based on natural language commands. A comprehensive ablation study examines the impact of trajectory definition and training data size on the model's performance. We demonstrate that even $2\%$ of the data helps get an average $mAP$ value of $30.6_{\pm{2.3}}$. Overall, this work contributes to enhancing the robot's capacity to comprehend and generalize tasks, improving human-robot interactions, and expanding the possibilities of robotics in various domains.

In future work, we will deploy a more efficient version of the proposed approach on an actual robot, leveraging smaller demonstrations based on the findings from the training data size ablation studies. This would involve learning a sub-task mapping from limited data, behavior cloning from language to sub-task trajectories, and exploring the combination of instructions to enhance the robot's learning capabilities and overall robustness.
\newpage
\bibliographystyle{plainnat}
\bibliography{references}


\onecolumn
\newpage
\section*{Appendix}
\subsection{Implementation Details}
The model is implemented in PyTorch. We set the hidden size $d = 256$, the number of layers in the encoder and decoder layer, $T=2$, and the number of possible changepoint is $N=10$. We use a dropout of $0.1$ for transformer layers and $0.5$ for input projection layers. We set the loss fusion hyper parameters as $\lambda_{L1}=10$, $\lambda_{iou}=1$, $\lambda_{cls}=4$, $\lambda_{s}=1$, $\Delta=0.2$. The model weights are initialized with Xavier init. We use AdamW with an initial learning rate of $1e-4$, and weight decay of $1e-4$ to optimize model parameters. The model is trained for $200$ epochs with a batch size of $256$. All training is conducted on a single NVIDIA GeForce RTX $3090$ GPU, with a training time of $4$ hours for a single iteration.

\subsection{Example}
\begin{figure*}[h!]
    \centering
    \includegraphics[scale=0.64]{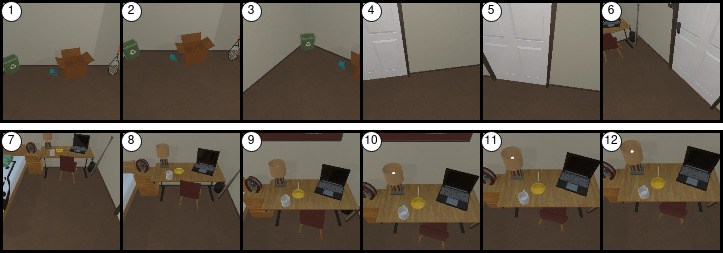}
    \caption{\textbf{[Sub Task 1]} \\\hspace{\textwidth}\textbf{Overall Goal}: Examine the vase by lamplight. \\\hspace{\textwidth}\textbf{Language Instruction}: Turn around and go to the desk. \\\hspace{\textwidth}\textbf{Discrete Actions to Image number mapping}: LookDown[1, 2], RotateLeft[3, 4, 5, 6, 7], MoveAhead[8, 9, 10], RotateLeft[11], LookDown[12]}
    \label{fig:init1o}
\end{figure*}

\begin{figure*}[h!]
    \centering
    \includegraphics[scale=0.64]{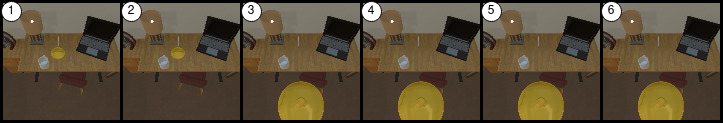}
    \caption{\textbf{[Sub Task 2]} \\\hspace{\textwidth}\textbf{Overall Goal}: Examine the vase by lamplight. \\\hspace{\textwidth}\textbf{Language Instruction}: Pick up the yellow vase. \\\hspace{\textwidth}\textbf{Discrete Actions to Image number mapping}: PickupObject[1, 2, 3, 4, 5, 6]}
    \label{fig:init2o}
\end{figure*}

\begin{figure*}[h!]
    \centering
    \includegraphics[scale=0.64]{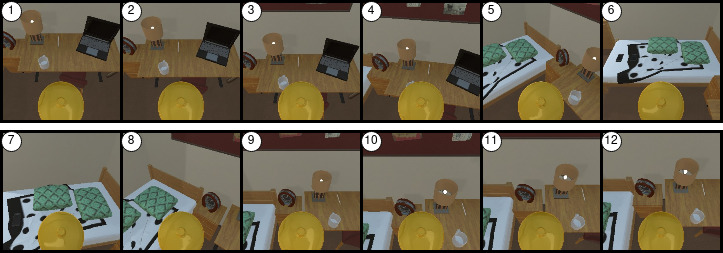}
    \caption{\textbf{[Sub Task 3]} \\\hspace{\textwidth}\textbf{Overall Goal}: Examine the vase by lamplight. \\\hspace{\textwidth}\textbf{Language Instruction}: Turn to the left, take a step, turn right to face to lamp. \\\hspace{\textwidth}\textbf{Discrete Actions to Image number mapping}: LookUp[1, 2], RotateLeft[3, 4, 5], MoveAhead[6, 7], RotateRight[7, 8, 9], MoveAhead[10, 11], LookDown[12]}
    \label{fig:init3o}
\end{figure*}

\begin{figure*}[h!]
    \centering
    \includegraphics[scale=0.64]{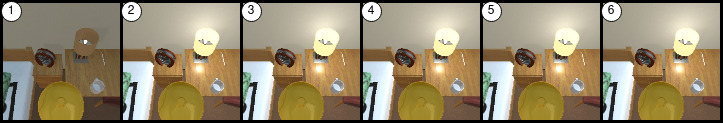}
    \caption{\textbf{[Sub Task 4]} \\\hspace{\textwidth}\textbf{Overall Goal}: Examine the vase by lamplight. \\\hspace{\textwidth}\textbf{Language Instruction}: Turn on the lamp on the desk. \\\hspace{\textwidth}\textbf{Discrete Actions to Image number mapping}: ToggleObjectOn[1, 2, 3, 4, 5, 6]}
    \label{fig:init4o}
\end{figure*}

\begin{figure*}[h!]
    \centering
    \includegraphics[scale=0.64]{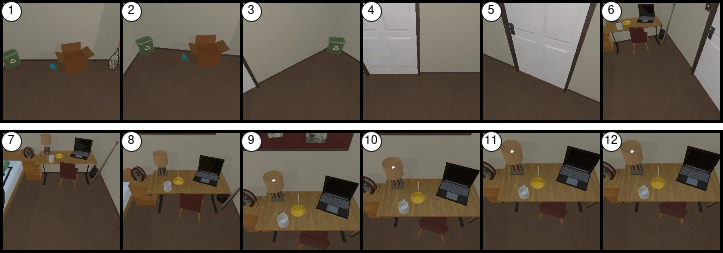}
    \caption{\textbf{Changepoint prediction} for Language Instruction (Sub Task 1 from above example): "Turn around and go to the desk" with a model trained on 100\% Training data}
    \label{fig:init100}
\end{figure*}

\begin{figure*}[h!]
    \centering
    \includegraphics[scale=0.65]{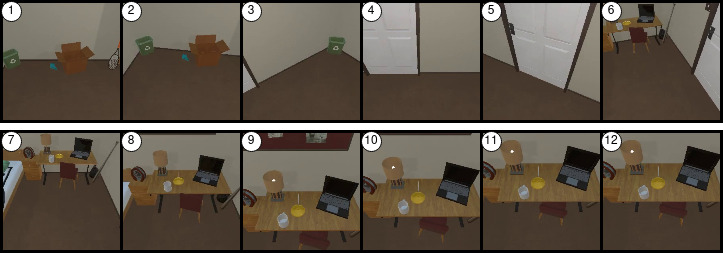}
    \caption{\textbf{Changepoint prediction} for Language Instruction (Sub Task 1 from above example): "Turn around and go to the desk" with a model trained on 50\% Training data}
    \label{fig:init50}
\end{figure*}

\begin{figure*}[h!]
    \centering
    \includegraphics[scale=0.65]{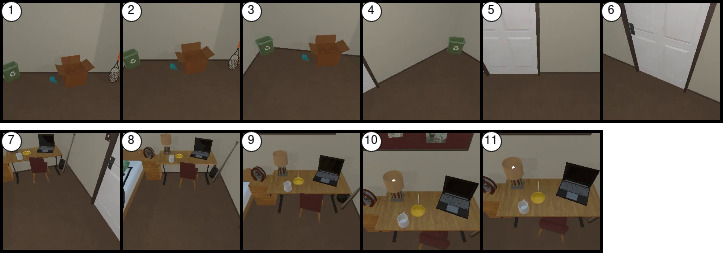}
    \caption{\textbf{Changepoint prediction} for Language Instruction (Sub Task 1 from above example): "Turn around and go to the desk" with a model trained on 25\% Training data}
    \label{fig:init25}
\end{figure*}

\begin{figure*}[h!]
    \centering
    \includegraphics[scale=0.65]{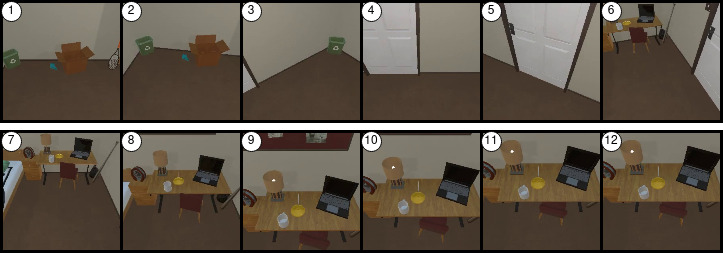}
    \caption{\textbf{Changepoint prediction} for Language Instruction (Sub Task 1 from above example): "Turn around and go to the desk" with a model trained on 10\% Training data}
    \label{fig:init10}
\end{figure*}

\begin{figure*}[h!]
    \centering
    \includegraphics[scale=0.65]{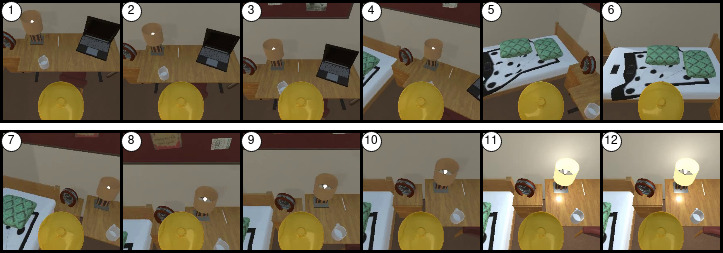}
    \caption{\textbf{Changepoint prediction} for Language Instruction (Sub Task 1 from above example): "Turn around and go to the desk" with a model trained on 2\% Training data}
    \label{fig:init2}
\end{figure*}

For a trajectory, whose overall goal is \textbf{Examine the vase by lamplight}, Figure[\ref{fig:init1o}, \ref{fig:init2o}, \ref{fig:init3o}, \ref{fig:init4o}] show the sub-segments of low-level language commands along with sub-tasks from trajectories consisting of images and discrete actions. Figure[\ref{fig:init100}, \ref{fig:init50}, \ref{fig:init25}, \ref{fig:init10}, \ref{fig:init2}] shows the prediction for changepoint detection for the sub-task in trajectory given the low-level natural language instruction "Turn around and go to the desk" by varying the training data points.

\begin{table*}[h!]
\begin{tabular}{ccccccc}
\hline
Dataset Split & \#Queries / \#Trajectory & \begin{tabular}[c]{@{}c@{}}Avg\\ Query Len\end{tabular} & \begin{tabular}[c]{@{}c@{}}Avg\\ Discrete Actions\end{tabular} & \begin{tabular}[c]{@{}c@{}}Avg\\ Instructions / Trajectory\end{tabular} & \begin{tabular}[c]{@{}c@{}}Avg Len (sec)\\ Moment / Trajectory\end{tabular} & \begin{tabular}[c]{@{}c@{}}Avg \\ Frames / Trajectory\end{tabular} \\ \hline
Train & 140.6K / 6.5K & 11 & 49.5 & 6.6 & 8.7 / 65.4 & 285.6 \\
Valid Seen & 5.5K / 250 & 11 & 49.6 & 6.6 & 8.8 / 67.7 & 285.2 \\ \hline
\end{tabular}
\caption{The Dataset Analysis after transforming the ALFRED data to changepoint detection data}
\label{tab:my-table}
\end{table*}

\begin{figure*}[h!]
    \centering
    \includegraphics[scale=0.64]{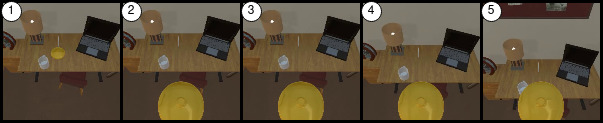}
    \caption{\textbf{Changepoint prediction} for Language Instruction (Sub Task 2 from above example): "Pick up the yellow vase." with a model trained on 100\% Training data}
    \label{fig:initii100}
\end{figure*}
\begin{figure*}[h!]
    \centering
    \includegraphics[scale=0.64]{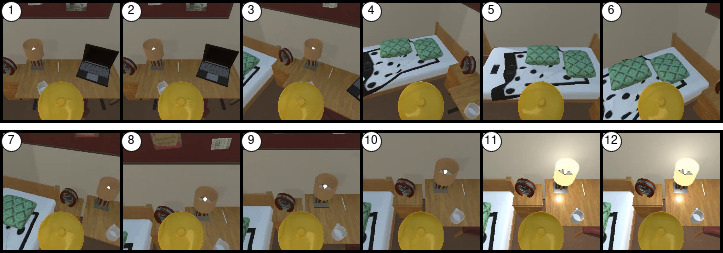}
    \caption{\textbf{Changepoint prediction} for Language Instruction (Sub Task 3 from above example): "Turn to the left, take a step, turn right to face to lamp." with a model trained on 100\% Training data}
    \label{fig:initiv100}
\end{figure*}
\begin{figure*}[h!]
    \centering
    \includegraphics[scale=0.64]{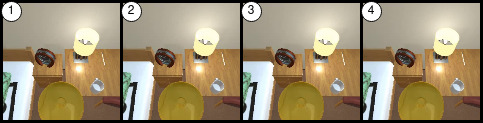}
    \caption{\textbf{Changepoint prediction} for Language Instruction (Sub Task 4 from above example): "Turn on the lamp on the desk" with a model trained on 100\% Training data}
    \label{fig:initiii100}
\end{figure*}


\end{document}